\documentclass[letterpaper, 10 pt, journal, twoside]{IEEEtran} 

\usepackage{tikz}
\usetikzlibrary{shapes, shapes.geometric, arrows, calc}
\usetikzlibrary{bayesnet}

\usepackage[ruled,vlined]{algorithm2e}
\SetKw{Continue}{continue}
\SetKw{Break}{break}

\usepackage{microtype}
\usepackage{graphicx}
\usepackage[font=small]{caption}
\usepackage{subcaption}

\usepackage{url}            
\usepackage{booktabs}       
\usepackage{tabularx}
\usepackage{amsfonts}       
\usepackage{nicefrac}       

\usepackage{amsmath,amsfonts,bm}
\usepackage{amssymb}

\usepackage{amsthm}
\usepackage{mathtools}
\usepackage{mathrsfs}

\usepackage{graphics} 
\usepackage{svg}
\usepackage{tikz}
\usepackage{pgfkeys}
\usepackage{standalone}
\usepackage{cite}
\usepackage{multirow}
\usepackage{threeparttable}
\usepackage{placeins}
\usepackage{adjustbox}
\usepackage[super]{nth}

\usepackage{hyperref}
\usepackage{siunitx}
\usepackage{cases}

\usepackage{acro}
\newcommand{\newac}[2]{\DeclareAcronym{#1}{short=#1,long=#2}}
\newac{CMB}    {Cosmic Microwave Background Radiation}
\newac{CNN}    {Convolutional Neural Network}
\newac{DEM}    {Digital Elevation Map}
\newac{ESA}    {European Space Agency}
\newac{GAN}    {Generative Adversarial Network} 
\newac{GNSS}   {Global Navigation Satellite System}
\newac{HDPR}   {Heavy Duty Planetary Rover}
\newac{IMU}    {Inertial Measurements Unit}
\newac{MSE}    {Mean Squared Error}
\newac{NLP}    {Natural Language Processing}
\newac{PConv}  {Partial convolutions}
\newac{PSNR}   {Peak signal-to-noise ratio}
\newac{ReLU}   {Rectified Linear Unit}
\newac{SLAM}   {Simultaneous Localization and Mapping}
\newac{SSIM}   {Structural Similarity Index Measure}
\newac{TV}     {Total Variation}
\newac{VAE}    {Variational Autoencoder}

\title{Reconstructing occluded Elevation Information in Terrain Maps with Self-supervised Learning}

\author{Maximilian Stölzle$^{1,2,3}$, Takahiro Miki$^{1}$, Levin Gerdes$^{2}$, Martin Azkarate$^{2}$, and Marco Hutter$^{1}$

\thanks{Manuscript received: September, 09, 2021; Revised December, 09, 2021; Accepted January, 05, 2022.}
\thanks{This paper was recommended for publication by Editor Javier Civera upon evaluation of the Associate Editor and Reviewers' comments.} 
\thanks{This work was partly supported by the PERASPERA activity of the European Commission.}
\thanks{$^{1}$Robotic Systems Lab, ETH Zürich, 8092 Zürich, Switzerland
        {\tt\scriptsize \{tamiki, mahutter\}@ethz.ch}}%
\thanks{$^{2}$Planetary Robotics Lab, European Space Agency, Keplerlaan 1, 2201 AZ Noordwijk, Netherlands
        {\tt\scriptsize \{levin.gerdes, martin.azkarate\}@esa.int}}%
\thanks{$^{3}$Cognitive Robotics, Delft University of Technology, Mekelweg 2, 2628 CD Delft, Netherlands
        {\tt\scriptsize M.W.Stolzle@tudelft.nl}}%
\thanks{Digital Object Identifier (DOI): see top of this page.}
}

\begin{document}
\bstctlcite{IEEEexample:BSTcontrol}

\maketitle


\begin{abstract}

Accurate and complete terrain maps enhance the awareness of autonomous robots and enable safe and optimal path planning.
Rocks and topography often create occlusions and lead to missing elevation information in the \ac{DEM}.
Currently, these occluded areas are either fully avoided during motion planning or the missing values in the elevation map are filled-in using traditional interpolation, diffusion or patch-matching techniques.
These methods cannot leverage the high-level terrain characteristics and the geometric constraints of line of sight we humans use intuitively to predict occluded areas.
We introduce a self-supervised learning approach capable of training on real-world data without a need for ground-truth information to reconstruct the occluded areas in the DEMs.
We accomplish this by adding artificial occlusion to the incomplete elevation maps constructed on a real robot by performing ray casting.
We first evaluate a supervised learning approach on synthetic data for which we have the full ground-truth available and subsequently move to several real-world datasets. These real-world datasets were recorded during exploration of both structured and unstructured terrain with a legged robot, and additionally in a planetary scenario on Lunar analogue terrain.
We state a significant improvement compared to the baseline methods both on synthetic terrain and for the real-world datasets.
Our neural network is able to run in real-time on both CPU and GPU with suitable sampling rates for autonomous ground robots.
We motivate the applicability of reconstructing occlusion in elevation maps with preliminary motion planning experiments.

\end{abstract}

\begin{IEEEkeywords}
Mapping, AI-enabled robotics
\end{IEEEkeywords}

\acresetall

\section{Introduction}\label{sec:introduction}
\IEEEPARstart{A}{s} we empower mobile robots to autonomously navigate to their goal, they rely on maps of the surrounding environment for traversability analysis and motion planning.
2.5D \acp{DEM} represent a memory-efficient and accurate approximation for most terrains as they project the 3D structure into two-dimensional grid cells.
Usually, robots use depth sensors such as LiDAR or stereo cameras to construct \acp{DEM}~\cite{Fankhauser2018ProbabilisticTerrainMapping}.
To enable safe and optimal path planning, we strive for complete and accurate elevation maps. 
In practice, our robots often encounter occlusions caused by terrain discontinuities such as rocks, obstacles or convex terrain characteristics which hide an area patch from the sensor's viewpoint.
Further, depth measurements can be degraded due to reflections, stereo matching failures, dust, or textureless surfaces which frequently lead to additional missing elevation information in the \ac{DEM}.
This motivates the need for solutions to reconstruct the missing elevation information in the terrain map.

We are inspired by the application of data-driven methods for inpainting of images~\cite{wang2020multistage, elharrouss2020image, demir2018patch} 
even for irregular holes~\cite{liu2018image} and aim to tailor these methods for the application of filling occlusion in terrain maps. 
Our method allows us to exploit prior information about the deployment area such as known terrain characteristics as well as the geometric constraints of line of sight for the reconstruction of missing elevation information using a neural network.
The method is based on a U-Net~\cite{ronneberger2015u} similarly to the state-of-the-art in neural network-based inpainting for vision problems~\cite{liu2018image, demir2018patch,
wang2020multistage, elharrouss2020image}.
We propose a new strategy for self-supervised learning applied to terrain maps based on adding artificial occlusion with ray casting.
As non-occluded ground-truth data is very difficult to acquire, this enables us to use incomplete real-world elevation maps for training our neural network.
We would like to point out that this self-supervised learning approach breaks ground towards a future where robots can actively improve their performance while they explore their environment in the line of the connected research on lifelong learning~\cite{thrun1995lifelong}.

\begin{figure*}[ht]
    \centering
    \includegraphics[width=1\textwidth]{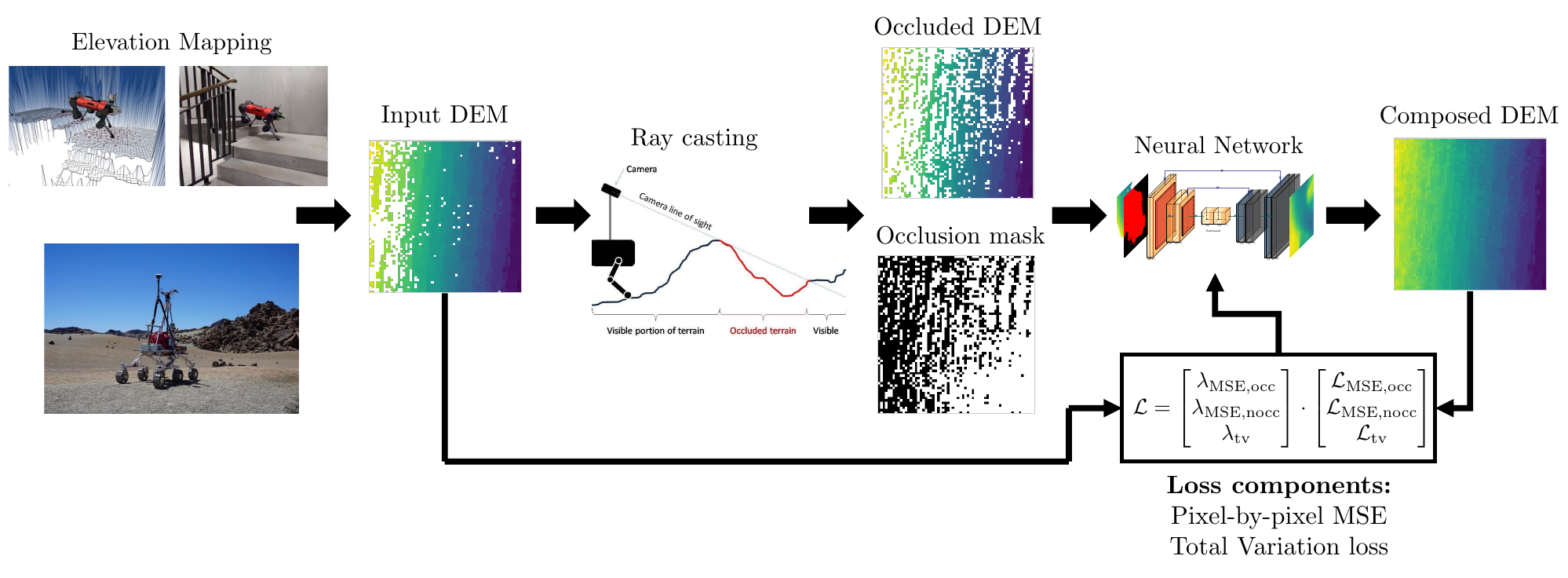}
    \caption{Self-supervised learning using artificial occlusion generated via ray casting for inpainting occlusion in terrain maps:
    partially occluded \ac{DEM} are sampled via robot-centric elevation mapping.
    Artificial occlusion is added by performing ray casting from a random vantage point which allows a neural network to be trained to reconstruct the \ac{DEM} without occlusion.
    The composed \ac{DEM} consists of patching the original occluded \ac{DEM} with the reconstruction in the occluded area. The ray casting graphic is adapted from Kolter et al.~\cite{kolter2009stereo}.}\label{fig:method}
    \vspace{-0.5cm}
\end{figure*}

Existing implementations to complete sparse terrain maps for mobile robots either rely on searching terrain patches with close resemblance in an offline library~\cite{kolter2009stereo} or very recently on neural networks trained with ground-truth data estimated offline~\cite{vsalansky2021pose}. The later approach~\cite{vsalansky2021pose} is based on the assumption that by \ac{SLAM} also having access to future measurements, the map will be more complete and accurate than the \ac{DEM} at the current time-step which is generated solely from current and past depth measurements. 
It should be noted that learned features will most likely be coupled with pose estimation errors and future flawed depth measurements through the \emph{ground-truth} map.

We conduct extensive bench-marking of both supervised and self-supervised learning approaches against various baseline methods. We evaluated on multiple synthetic terrain types as well as real-world datasets collected with the legged robot ANYmal~\cite{hutter2016anymal} and the planetary rover test-bed HDPR~\cite{boukas2016hdpr}.
We state a decrease between $\SI{52}{\percent}$ and \SI{82}{\percent} in \ac{MSE} error with our self-supervised method on the real-world datasets compared to the respective best performing baseline approach.
Finally, we observe in preliminary motion planning experiments that using reconstructed \acp{DEM} paths can be planned further ahead compared to planning with incomplete elevation maps. Consequently, the robot needs to stop less and avoids having to wait for re-planning of the path to finish.

We would like to emphasize the following contributions of this paper: 
We propose a self-supervised learning method for filling-in sparse \acp{DEM} by adding artificial occlusion with ray casting enabling training on partially occluded real-world datasets. We introduce an iterative algorithm for realistic occluded map data generation and to guide the sampling of vantage points for ray casting towards occlusion masks with suitable amounts of occlusion.


\section{Related Work}\label{sec:related_work}
\textbf{Image inpainting:}
Inpainting techniques utilise the structure and texture of an image to fill-in missing parts and can be used to for example remove unwanted objects from the image~\cite{Mehra2020FromTI}.
Traditional image inpainting methods can be separated into two categories: 
While diffusion-based inpainting methods leverage isophote lines to continuously propagate high-order derivatives of local pixel density from the exterior into the hole of the image~\cite{ebrahimi2013navier}, 
patch-based methods such as PatchMatch~\cite{barnes2009patchmatch} search the known regions of the image at the patch-level and copy the best-matching patches into the missing region.
%
Nowadays, data-driven learning-based approaches are the state-of-the-art for image inpainting \cite{liu2018image, wang2020multistage, elharrouss2020image, demir2018patch}.
Most of them rely on an adoption of the U-Net~\cite{ronneberger2015u} model architecture.
Many recent papers~\cite{liu2018image, song2018contextual, wang2020multistage} include perceptual and style losses~\cite{gatys2015neural}.
They rely on high-level features extracted using a pretrained VGG-16 
network. 
These perceptual and style losses work similarly to adversarial losses by not just enforcing convergence of the reconstruction on a pixel-by-pixel basis, but also enforcing encoded high-level characteristics to be similar.
Partial convolutions can be appropriate for irregular holes as convolutional strategies might fail to adjust to the changing shapes of the hole boundaries~\cite{liu2018image}. 
A \ac{GAN}'s ability to generate high-fidelity images motivates the use of adversarial methods to produce sharper inpaintings~\cite{qiu2019void, wang2020multistage, demir2018patch}.
\\
\textbf{Self-supervised learning:}
Self-supervised learning is essential to alleviate the gap between synthetic data and real-world data (Syn2Real) and enable training on real-world datasets for which complete ground-truth information is challenging to acquire.
Zhan et al.~\cite{Zhan_2020_CVPR} propose a self-supervised approach for de-occlusion in the framework of scene understanding by overlaying semantically-extracted objects onto images.
Dai et al.~\cite{dai2020sg} uses a self-supervised learning approach for scene completion of sparse real-world RGB-D scans using a \ac{GAN} by randomly removing some of the scans.
\\
\textbf{Elevation mapping:}
Robot-centric elevation mapping uses pose estimates from \ac{IMU} or odometry and local distance measurements from a laser range sensor, structured light, or stereo camera to derive a local 2.5D elevation map of the environment with the robot centered in the grid~\cite{Fankhauser2018ProbabilisticTerrainMapping}.
The uncertainty of elevation values is predicted through sensor noise models and localization drift estimates and can be very valuable for motion planning~\cite{Fankhauser2018ProbabilisticTerrainMapping}.
\\
\textbf{Completing sparse elevation maps:}
Relatively few works consider the challenge of completing sparse robotic elevation maps. 
One early work by Kolter et al.~\cite{kolter2009stereo} proposes a non-parametric algorithm based on texture synthesis methods to fill-in the missing terrain portions. This method is comparable to patch-based methods in image inpainting as it searches for patches in a library that closely resemble the partially occluded terrain patch with the additional enforcement of geometric constraints with respect to line of sight for the posterior distribution over the missing region.
In recent years, data-driven methods were explored for completing sparse \acp{DEM}.
Qiu et al.~\cite{qiu2019void} leverage a deep convolutional \ac{GAN} to fill the voids of large-scale geospatial data \ac{DEM} with the loss function being composed of pixel-wise, contextual and perceptual constraints.
For mobile robots however, there is even more information available than just map reconstruction losses as the robot pose is estimated and recorded throughout the trajectory~\cite{vsalansky2021pose}. The robot pose hints at a likely local terrain shape which can be leveraged as part of a weakly-supervised KKT-loss term.
This is in particular relevant for partially flexible terrain for which lidar measurements over-estimate the support height.
The supervision through recorded poses is extended by training a pose regressor leading to a pose prediction-loss for areas of the terrain map not traversed by the robot.
The reconstruction loss itself is trained by estimating a ground-truth map offline which also includes future measurements and thus is more complete and likely accurate than the current \ac{DEM}~\cite{vsalansky2021pose}.
The confidence in occupancy of parts of the map can furthermore be used to actively guide the depth measurement sensor to explore sparse areas of the map~\cite{zimmermann2017learning}.

\section{Methodology}\label{sec:methodology}
First, we give a brief overview of our method as visualized in Fig.~\ref{fig:method}: we consider sparse 2.5D elevation maps and strive to fill-in the missing elevation information to match the ground-truth \ac{DEM} as accurately as possible. 
We propose to use an U-Net\cite{ronneberger2015u}-like neural network to inpaint the occluded 2.5D elevation map with the inputs consisting of the \ac{DEM} and a binary occlusion mask. Supervised learning requires us to know ground-truth data to compute a training loss and after back-propagation optimize the neural network weights. 
As complete and accurate ground-truth information is rarely available for real-world datasets, we introduce a self-supervised method which leverages ray casting to further occlude the \ac{DEM} given by the robot sensors.
This allows us to compute a \ac{MSE} and \ac{TV} loss between the occluded \ac{DEM} and the less occluded input \ac{DEM}.

\begin{figure*}[ht]
  \centering
  \subfloat{\includegraphics[width=0.5\textwidth]{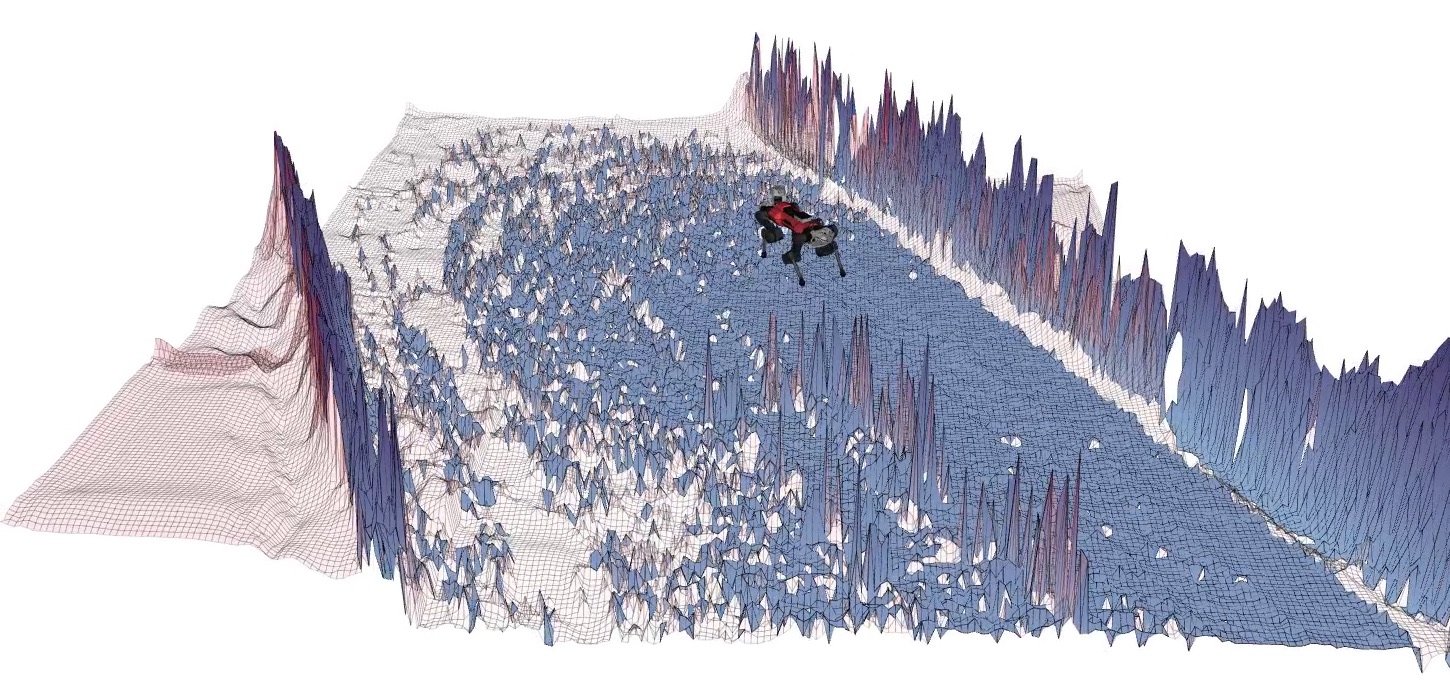}\label{fig:snap_gonzen_1}}
  \subfloat{\includegraphics[width=0.5\textwidth]{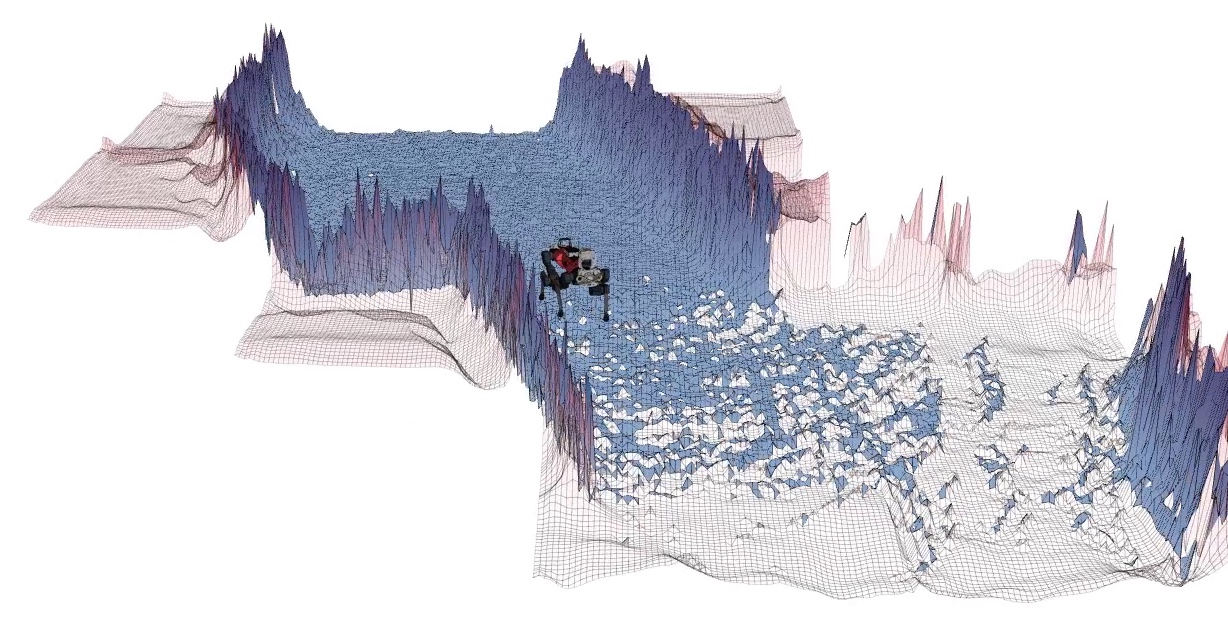}\label{fig:snap_gonzen_2}}
  \caption{Snapshots of inference using the trained neural network on the Gonzen mine dataset recorded with the ANYmal legged robot. The occluded \ac{DEM} is marked in a color scale from white to blue (lowest to highest elevation), while the reconstruction is visualized from white to red with a slight opacity. To improve the clarity of presentation, we only inpaint subgrids which are less than \SI{85}{\percent} occluded as the neural network requires sufficient input elevation information.}
  \vspace{-0.6cm}
\end{figure*}
\subsection{Problem statement}\label{sec:methodology_problem_statement}
We consider a 2.5D \ac{DEM} $\mathbf{m}_\mathrm{occ} \in \mathcal{R}^{n \times m}$ of a terrain patch.
The terrain patch of length $l$ and width $w$ is discretized at a resolution $r$ to compose a regular grid of dimension $n \times m$, which encodes the elevation at every grid cell.
However, some elevation information is missing because of geometrical occlusion or other reasons such as degraded depth measurements.
We create a binary occlusion mask $\mathbf{M}_\mathrm{occ} \in [0; 1]^{n \times m}$ which assigns a value of 1 to grid cells with missing elevation information and 0 to grid cells with known elevations.
We define our problem as such, that we want to estimate a reconstructed \ac{DEM} $\mathbf{m}_\mathrm{rec}$ which matches the ground-truth \ac{DEM} $\mathbf{m}_\mathrm{gt}$ as closely as possible.
During inference, we construct a composition $\mathbf{m}_{\mathrm{comp}}$ between the original input $\mathbf{m}_\mathrm{occ}$ for the non-occluded grid cells and the reconstruction $\mathbf{m}_\mathrm{rec}$ for the occluded grid cells.

\subsection{Ray casting}\label{sub:methodology_raycasting}
We use ray casting to compute occluded \acp{DEM} based on the synthetic ground-truth \acp{DEM} in addition to generating artificial occlusion in the framework of self-supervised learning.
We developed a lightweight C\texttt{++} component 
to perform fast ray casting of an entire grid map from a given vantage point. 
The ray casting algorithm takes a DEM and a vantage point $\mathbf{x}_v$ as inputs and then
iterates through every cell in the grid and checks for each cell whether the cell is visible from the vantage point.

\subsection{Self-supervised learning}\label{sub:methodology_self_supervised}
We propose a self-supervised learning methodology to train on real-world datasets without ground-truth to learn to fill-in missing elevation information of DEMs.
We reformulate the training setup by using our partly occluded real-world elevation map as a target and creating additional artificial occlusion which is used as an input into the neural network.
We considered a dilation of the already occluded area in addition to randomly occluding pixels. However, this approach does not render realistic and diverse occlusion masks.
We decided to add artificial occlusion by ray casting from a randomly sampled vantage point.

We employ an iterative algorithm to generate useful artificial occlusion masks with occlusion ratios between \SI{0.1}{\percent} and \SI{50}{\percent} and choose the elevation offset $o$ of the vantage point accordingly.
First, we sample a random vantage point from the grid with a uniform distribution.
Then after a random elevation offset for the vantage point $o \sim \mathcal{U}\left ( o_\text{min}, o_\text{max} \right )$ is sampled, we perform ray casting for the chosen vantage point and evaluate the resulting occlusion ratio (e.g., number of occluded grid cells over number of total grid cells).

We require that the occlusion ratio $r_\text{occ}$ lies within the interval $[r_\text{occ,min}, r_\mathrm{occ,max}]$.
This threshold interval was tuned in a selection study and set to $[\SI{0.1}{\percent}, \SI{50}{\percent}]$.
If the occlusion ratio does not lie within the interval, we sample a new elevation offset.
This time, we adjust the range of the uniform distribution: if $r_\mathrm{occ} > r_\mathrm{occ,max}$, we increase $o_\text{min}$ to the previously sampled elevation offset $o$, and if $r_\text{occ} < r_\text{occ,min}$, we decrease $o_\mathrm{max}$ to the previously sampled elevation offset $o$.
If $\left\lVert o_\text{max} - o_\text{min} \right\rVert < \SI{0.05}{m}$, we enforce a minimal sampling range by respectively decreasing $o_\text{min}$ by \SI{0.05}{m} or increasing $o_\text{max}$ by \SI{0.05}{m}.
We repeat this algorithm until either we have satisfied our occlusion ratio constraint $[r_\text{occ,min}, r_\text{occ,max}]$ or we have reached the maximum number of iterations (15).
We reformulate our loss function to only consider the areas of artificial occlusion and ignore the already occluded areas in the target DEM. 

\subsection{Model}\label{sub:methodology_model}
We adopt an U-Net~\cite{ronneberger2015u} as the architecture of our neural network which is often used in the literature for image and video inpainting~\cite{liu2018image, wang2020multistage}. 
The input is composed of two channels: the occluded elevation map $\mathbf{m}_\mathrm{occ}$ and the binary occlusion mask $\mathbf{M}_\mathrm{occ}$.
As the occluded \ac{DEM} contains missing elevation values represented as NaNs computationally, we replace them with a floating point number by linearly interpolating for synthetic datasets and with the constant $0.0$ for real-world datasets after a selection study. A reason could be that linear interpolation is performing worse for the higher occlusion ratios of the real-world datasets.
We also implement input and output normalization: we compute the mean of the non-occluded elevation values of a \ac{DEM} and subtract this mean from each elevation value in the grid.
After receiving the output of the model, we add this mean back to each elevation value before computing the loss. 
We adapt our model from the vanilla U-Net architecture and make slight adjustments to the number of max-pooling steps because we are working with smaller input images (64x64px instead of 572x572px) than in the original paper~\cite{ronneberger2015u} and we want to keep the network as lightweight as possible.
Thus, we limit the number of max-pooling operations to 3 (instead of 5 in the original paper) and treat the number of channels in each hidden dimension as a hyperparameter for which we select 64, 128, and 256 channels for our hidden dimensions. 

\subsection{Loss function}\label{sub:methodology_loss_function}
We enforce a \ac{MSE} loss between the ground-truth \ac{DEM} $\mathbf{m}_{\mathrm{gt}}$ and the reconstructed \ac{DEM} $\mathbf{m}_{\mathrm{rec}}$ as our pixel-by-pixel reconstruction loss.
We separate the \ac{MSE} loss for the occluded region of the DEM $MSE_{\text{occ}}$ and the non-occluded region of the DEM $MSE_{\text{nocc}}$.
A \ac{TV} loss $\mathcal{L}_\mathrm{tv}$~\cite{johnson2016perceptual} has shown to be valuable as a smoothing penalty on the reconstruction of the occluded area. 
We compute the final loss as a weighted sum of all loss components with weights of $1$ and $10$ for the \ac{MSE} loss of the non-occluded and occluded area respectively, 
and scale of $0.1$ for the \ac{TV} loss.

\subsection{Training}\label{sub:methodology_training}
We base our software to train the models on PyTorch~\cite{NEURIPS2019_9015}.
After every training epoch on the training set, a pixel-by-pixel \ac{MSE} loss is evaluated for the occluded area $\mathcal{L}_\text{MSE,occ}$ as our validation loss.
The training is stopped if either a maximum number of epochs is reached (100) or the validation loss did not improve during a specified number of epochs (50).
We use the Adam~\cite{kingma2014adam} optimizer with a learning rate of $0.0001$, a weight decay of $0.001$, and the beta coefficients $(0.9, 0.999$).

\section{Experiments and Results}\label{sec:results}
We evaluate our method both quantitatively and qualitatively on multiple synthetic and real-world datasets. More specifically, we compare the performance of the self-supervised learning approach to several trivial baselines such as linear~\cite{nouza2014safe} and cubic interpolation and traditional inpainting methods with hand-tuned heuristics such as Navier-Stokes~\cite{ebrahimi2013navier} and Telea~\cite{telea2004image}.

\begin{table*}
\centering
\caption{Results for synthetic datasets reporting the reconstruction performance for the occluded area averaged over three random seeds.}
\vspace{0.25cm}
\begin{scriptsize}
\begin{tabular}{lllllll}\toprule
\textbf{Method} & \textbf{Terrain} & $\ell_{1,\text{rec},\text{occ}}$ & $MSE_{\text{rec},\text{occ}}$ & $PSNR_\text{rec,occ}$ & $SSIM_\text{rec}$ & $SSIM_\text{comp}$\\
\midrule
Linear interpolation~\cite{nouza2014safe} & \multirow{6}{*}{Hills} & $0.0123$ & $0.00052$ & $37.95$ & $0.99261$ & $0.99261$\\
Cubic interpolation &  & $\mathbf{0.0103}$ & $0.00046$ & $38.54$ & $\mathbf{0.99326}$ & $0.99326$\\
Telea~\cite{telea2004image} & & $0.0231$ & $0.00223$ & $31.65$ & $0.98686$ & $0.98686$\\
Navier-Stokes~\cite{ebrahimi2013navier} &  & $0.0226$ & $0.00302$ & $30.33$ & $0.98882$ & $0.98882$\\
Supervised &  & $0.0121 \pm 0.0005$ & $\mathbf{0.00033 \pm 0.00001}$ & $\mathbf{40.0 \pm 0.2}$ & $0.9888 \pm 0.0004$ & $\mathbf{0.9950 \pm 0.0002}$\\
Self-supervised & & $0.0116 \pm 0.0003$ & $0.00035 \pm 0.00001$ & $39.7 \pm 0.2$ & $0.987 \pm 0.001$ & $0.9949 \pm 0.0003$\\
\midrule
Linear interpolation~\cite{nouza2014safe} & \multirow{6}{*}{Stairs} & $0.0534$ & $0.00472$ & $32.43$ & $0.97273$ & $0.97273$\\
Cubic interpolation &  & $0.0633$ & $0.00937$ & $29.44$ & $0.95952$ & $0.95952$\\
Telea~\cite{telea2004image} & & $0.0608$ & $0.00752$ & $30.40$ & $0.95543$ & $0.95538$\\
Navier-Stokes~\cite{ebrahimi2013navier} &  & $0.0601$ & $0.00877$ & $29.73$ & $0.95399$ & $0.95399$\\
\textbf{Supervised} &  & $0.012 \pm 0.003$ & $\mathbf{0.0003 \pm 0.0001}$ & $\mathbf{44 \pm 1}$ & $\mathbf{0.992 \pm 0.002}$ & $0.9982 \pm 0.0005$\\
Self-supervised &  & $0.013 \pm 0.002$ & $0.0005 \pm 0.0001$ & $42.2 \pm 0.8$ & $0.992 \pm 0.002$ & $0.9980 \pm 0.0003$\\
\midrule
Linear interpolation~\cite{nouza2014safe} & \multirow{6}{*}{Random Boxes} & $0.0660$ & $0.00721$ & $09.35$ & $0.78321$ & $0.78321$\\
Cubic interpolation &  & $0.0674$ & $0.00784$ & $08.98$ & $0.77970$ & $0.77970$\\
Telea~\cite{telea2004image} & & $0.0678$ & $0.00794$ & $08.93$ & $0.7784$ & $0.7784$\\
Navier-Stokes~\cite{ebrahimi2013navier} &  & $0.0655$ & $0.00761$ & $09.12$ & $0.7806$ & $0.7806$\\
\textbf{Supervised} &  & $\mathbf{0.0118 \pm 0.0009}$ & $\mathbf{0.00078 \pm 0.00003}$ & $\mathbf{19.0 \pm 0.1}$ & $\mathbf{0.88 \pm 0.03}$ & $\mathbf{0.959 \pm 0.004}$\\
Self-supervised &  & $0.0151 \pm 0.0007$ & $0.0011 \pm 0.0001$ & $17.7 \pm 0.5$ & $0.72 \pm 0.09$ & $0.949 \pm 0.006$\\
\bottomrule
\end{tabular}
\end{scriptsize}
\vspace{-0.5cm}
\label{tab:results_synthetic_datasets}
\end{table*}

\subsection{Datasets}\label{sub:datasets}
\textbf{Synthetic dataset:}
We evaluate our methods on synthetic datasets using three different terrain types: hills, standard stairs, and random boxes. 
This allows us to train with supervised learning but also to benchmark our methods against an accurate ground-truth, as it is usually very hard to acquire a matching ground-truth for a large dataset of real-world occluded terrain maps.
We generate terrain maps of size 64x64px and with a resolution of $\SI{0.04}{m \per px}$.
We attribute 25,000 samples to the training, and 2,500 samples each to the validation and test set.
The hills terrain is generated using a distribution of Perlin noise~\cite{lagae2010survey}. 
We also implement terrains with highly regular patterns in standard stairs and random boxes.
While the standard stairs terrain is generated by stacking boxes in a uniform direction with a regular step size using fixed box dimension, boxes are stacked at various angles with randomly sampled dimensions for the random boxes terrain.
After spawning the terrain map for the chosen terrain type, we perform ray casting to gather the occluded elevation map $\mathbf{m}_\mathrm{occ}$ and the binary occlusion mask $\mathbf{M}_\mathrm{occ}$. 
We sample a random robot position from a uniform distribution $x \sim \mathcal{U}\left(\SI{-1.25}{m}, \SI{1.25}{m} \right)$ to perform ray casting. 
We subsequently infer the elevation of the robot position from the \ac{DEM} at the corresponding pixel and sample a random elevation offset of the vantage point for ray casting $o_\mathrm{vp} \sim \mathcal{U}\left(\SI{0.2}{m}, \SI{0.5}{m} \right)$ for the hills terrain and $o_\mathrm{vp} \sim \mathcal{U}\left(\SI{0.2}{m}, \SI{0.3}{m} \right)$ for the standard stairs and random boxes terrains respectively.
\\
\textbf{ANYmal datasets:}
We consider datasets recorded using the ANYmal~\cite{hutter2017anymal} legged robot in three different terrains. The ANYmal robot is equipped with a dome LiDAR sensor and an \ac{IMU} to perceive 3D point clouds of its environment and odometry information respectively.
2.5D \acp{DEM} of size 300x300px and resolution \SI{4}{cm} are derived using robot-centric elevation mapping~\cite{Fankhauser2014RobotCentricElevationMapping, Fankhauser2018ProbabilisticTerrainMapping}.
The terrains include several datasets recorded while the robot is traversing on a staircase and on an obstacle course at ETH Zürich, and one dataset capturing the subterranean exploration of the Gonzen mine in Switzerland~\cite{tranzatto2021cerberus}.
For the ANYmal datasets, we use approximately \SI{80}{\percent} of the subgrid \acp{DEM} for the training set and \SI{10}{\percent} each for the validation and test sets respectively.
The ETH stairs, ETH obstacles course, and Gonzen mine datasets contain in total 26,233, 37,274 and 16,459 samples.
We divide the 300x300px \acp{DEM} into 16 subgrids à 75x75px each, to increase generalization capabilities, reduce the GPU memory requirements and exclude empty parts.
We uploaded a video of the inference of our method on the Gonzen mine dataset~\cite{tranzatto2021cerberus} to YouTube\footnote{\url{https://youtu.be/2Khxeto62LQ}}.
\\
\textbf{Tenerife Lunar Analogue dataset:}
We evaluate our methods on a lunar analogue dataset which was collected during a field test campaign in June 2017 at Minas de San José on the Teide Volcano on the island of Tenerife using the \ac{HDPR}~\cite{boukas2016hdpr}, 
a lab rover testbed with resemblance to the Rosalind Franklin rover \cite{vago2017habitability} 
flying to Mars in 2022 during the ExoMars mission~\cite{azkarate2020gnc}.
The dataset contains \ac{GNSS} recordings, 
the rovers raw inertial data, and images from three stereo cameras captured during several traverses over the duration of multiple days including a variety of lighting conditions and different pre-planned paths.
We choose elevation maps generated during a traverse on an isolated side track as our test set because they are completely unseen during all other traverses.
We apply the GA SLAM~\cite{geromichalos2020gaslam} technique to extract DEMs from the raw dataset.
As we want to minimize all errors in the local maps introduced by drifts in the localization, we use the ground-truth robot pose provided by the \ac{GNSS} as opposed to the visual odometry from the LocCam to translate the local map to the next time step and rotate the incoming point cloud before stereo processing.
We only select keyframe elevation maps with a \ac{PSNR} smaller than \SI{50}{dB} between the current and the last accepted sample which drastically decreases the computational load during training and also prevents overfitting issues which would otherwise arise from having too many similar samples.
We divide the 600x600px \ac{DEM}s into 64 subgrids à 75x75px.
This policy results in 42,600 samples for the training set, 1,000 samples for the validation set and 7,950 samples for the test set.

\subsection{Evaluation metrics}\label{sub:methodology_evaluation_metrics}
Similar to other publications on the topic of image inpainting such as \cite{liu2018image, yu2018generative}, 
we state the $\ell_1$ loss, the \ac{PSNR} and the \ac{SSIM}~\cite{wang2004image}.
We use the separated test set for all evaluation results.

The \ac{PSNR} is a function of the total \ac{MSE} loss of the occluded area and the maximum dynamic range of the grid-values $L$ determined by computing the delta between the highest and lowest elevation of the ground-truth \acp{DEM} 
of the entire test dataset. 
We only report the evaluation metrics for the occluded area as we argue that a composed DEM incorporating the non-occluded parts of the input DEM $\mathbf{m}_\mathrm{comp}$ can be easily created and thus the reconstruction capability of the non-occluded area by the model is not essential.

We compute the \ac{SSIM} between the ground-truth \ac{DEM} $\mathbf{m}_{\mathrm{gt}}$ and the reconstructed DEM $\mathbf{m}_{\mathrm{rec}}$ as $SSIM_\text{rec}$ and similarly between the ground-truth and the composed \ac{DEM} $\mathbf{m}_{\mathrm{comp}}$ as $SSIM_\text{comp}$.
The metric is quantified over the entire \ac{DEM} and not just the occluded region as it is not defined for irregular patches.
We use the constants $C_1 = (k_1 L)^2$, $C_2 = (k_2 L)^2$ and $C_3 = C_2 / 2$, where $L$ signifies the dynamic range as previously used for the \ac{PSNR},
and $k_1 = 0.01$ and $k_2 = 0.03$. We calculate the average $\mu$ and the variance $\sigma^2$ with a Gaussian filter with a 1D kernel of size $11$ and sigma $1.5$ for every \ac{DEM} separately. 
We set the weight exponents $\alpha$, $\beta$, and $\gamma$ 
all to one. 
The \ac{SSIM} metric can only be evaluated on complete \acp{DEM} and thus requires knowledge of the entire ground-truth.

\begingroup
\setlength{\tabcolsep}{2.5pt} 
\begin{table}\scriptsize
\centering
\caption{Results for real-world datasets evaluated using artificial occlusion generated with ray casting on the test set averaged over five random seeds for the baseline methods linear- / cubic interpolation, Telea~\cite{telea2004image} and Navier-Stokes~\cite{ebrahimi2013navier}, and self-supervised learning. The chosen unit of elevation is meters.}
\vspace{0.25cm}
\begin{tabular}{lllll}\toprule
\textbf{Method} & \textbf{Terrain} & $\ell_{1,\text{occ}}$ & $MSE_{\text{occ}}$ & $PSNR_{occ}$\\
\midrule
Linear & \multirow{5}{*}{ETH St.} & $0.1014$ & $0.05544$ & $25.98$\\
Cubic & & $0.1193$ & $0.08715$ & $24.02$\\
Telea & & $0.0958$ & $0.04980$ & $26.45$\\
Navier-St. &  & $0.0956$ & $0.05339$ & $26.14$\\
\textbf{Self-super.} &  & $\bf{0.071 \pm 0.004}$ & $\bf{0.018 \pm 0.001}$ & $\bf{30.9 \pm 0.3}$\\
\midrule
Linear & \multirow{5}{*}{Obstacles} & $0.2981$ & $0.34646$ & $23.77$\\
Cubic & & $0.3453$ & $0.54131$ & $21.83$\\
Telea & & $0.2873$ & $0.35048$ & $23.72$\\
Navier-St. &  & $0.2801$ & $0.34751$ & $23.76$\\
\textbf{Self-super.} &  & $\bf{0.173 \pm 0.002}$ & $\bf{0.166 \pm 0.003}$ & $\bf{26.97 \pm 0.07}$\\
\midrule
Linear & \multirow{5}{*}{Gonzen} & $0.1977$ & $0.14473$ & $14.87$\\
Cubic & & $0.2242$ & $0.20689$ & $13.32$\\
Telea & & $0.2035$ & $0.16535$ & $14.29$\\
Navier-St. &  & $0.2020$ & $0.16787$ & $14.23$\\
\textbf{Self-super.} &  & $\bf{0.081 \pm 0.002}$ & $\bf{0.026 \pm 0.001}$ & $\bf{22.3 \pm 0.2}$\\
\midrule
Linear & \multirow{5}{*}{Tenerife} & $0.0881$ & $0.02577$ & $40.81$\\
Cubic & & $0.0929$ & $0.03106$ & $40.00$\\
Telea & & $0.0898$ & $0.02714$ & $40.58$\\
Navier-St. &  & $0.0881$ & $0.02705$ & $40.60$\\
\textbf{Self-super.} &  & $\bf{0.0502 \pm 0.0009}$ & $\bf{0.0110 \pm 0.0002}$ & $\bf{44.51 \pm 0.09}$\\
\bottomrule
\end{tabular}
\label{tab:results_real_world_datasets}
\vspace{-0.5cm}
\end{table}
\endgroup

\begin{figure*}[ht]
\centering
\includegraphics[width=1\textwidth]{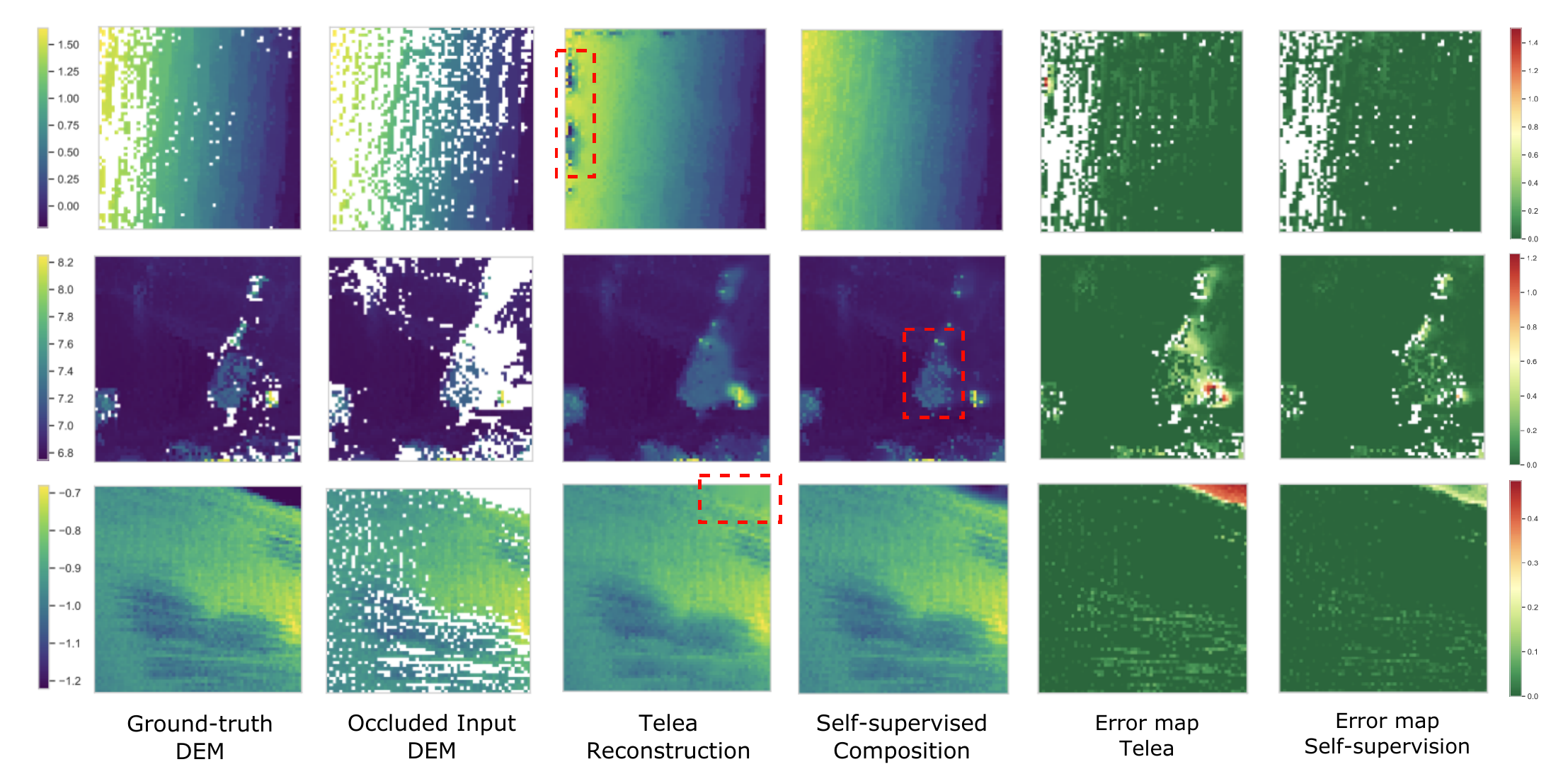}
\caption{Qualitative samples comparing the Telea~\cite{telea2004image} baseline method and self-supervised learning on various real-world test sets. The ground-truth \ac{DEM} describes the local, partially occluded terrain map subgrid sampled by the robot. We add artificial occlusion with ray casting to derive the occluded \ac{DEM} which serves as the input to both inpainting methods. We visualize the error compared to the partially occluded ground-truth map for Telea and our method. All units are in meter. \textbf{First row - ETH stairs:} the Telea reconstruction is blurring the stairs and additionally inpainting two holes with lower elevation values on the left side. \textbf{Second row - Gonzen mine:} our method reconstructs the obstacle in the center with higher accuracy leveraging knowledge about the terrain characteristics. \textbf{Third row - Tenerife lunar analogue:} self-supervised learning respects the geometric constraints of line of sight and reconstruct the occluded area in the top-right corner to have a lower elevation.}\label{fig:qualitative_samples}
\vspace{-0.5cm}
\end{figure*}

\subsection{Experiments on synthetic datasets}\label{sub:experiments_synthetic_datasets}
We evaluate both baseline methods, supervised and self-supervised approaches on the three synthetic datasets and report 
the results in Table~\ref{tab:results_synthetic_datasets} including the the \ac{SSIM} for both the reconstructed and the composed \ac{DEM}.
We train and evaluate experiments on three different random seeds and report a confidence interval with mean and standard deviation for those experiments.
For supervised learning, we directly use the non-occluded ground-truth as our training targets, while for self-supervised learning we adopt the occluded \ac{DEM} from the dataset as our training target and add artificial occlusion using ray casting. 

We show \SI{24}{\percent}, \SI{90}{\percent}, and \SI{85}{\percent} reductions in \ac{MSE} error for the occluded area compared best performing baseline approach on each dataset using self-supervised learning based on generating artificial occlusion with ray casting for the synthetic hills, standard stairs, and random boxes terrains respectively.
Substantial improvements can also be noted for the $\ell_1$ and \ac{SSIM} error metrics.
We would like to point out larger improvements for highly-structured terrain such as standard stairs and random boxes compared to the baseline methods.
Surprisingly, the performance of self-supervised learning is only slightly worse (approx. \SI{0.3}{dB}-\SI{2.3}{dB} in \ac{PSNR}) compared to fully supervised learning.
While (cubic) interpolation baseline methods perform very well on the smooth hills terrain, their performance is much worse on more structured terrain and similar to Telea~\cite{telea2004image} and Navier-Stokes~\cite{ebrahimi2013navier}.

\begin{figure*}[ht]
  \centering
  \subfloat{\includegraphics[width=0.495\textwidth]{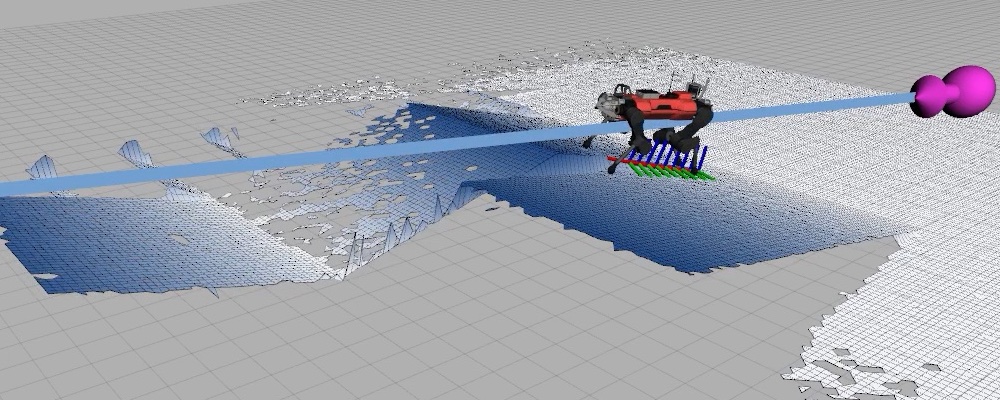}}
  \hfill
  \subfloat{\includegraphics[width=0.495\textwidth]{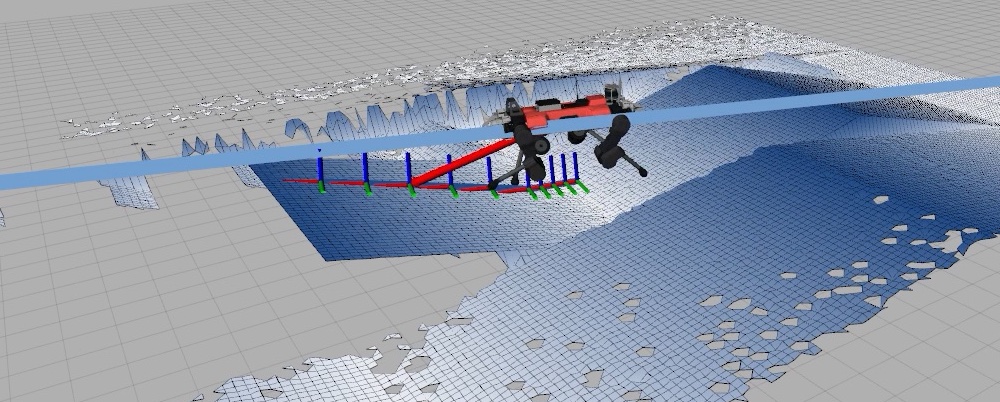}}
  \vspace{0.15cm}\\
  \subfloat{\includegraphics[width=0.495\textwidth]{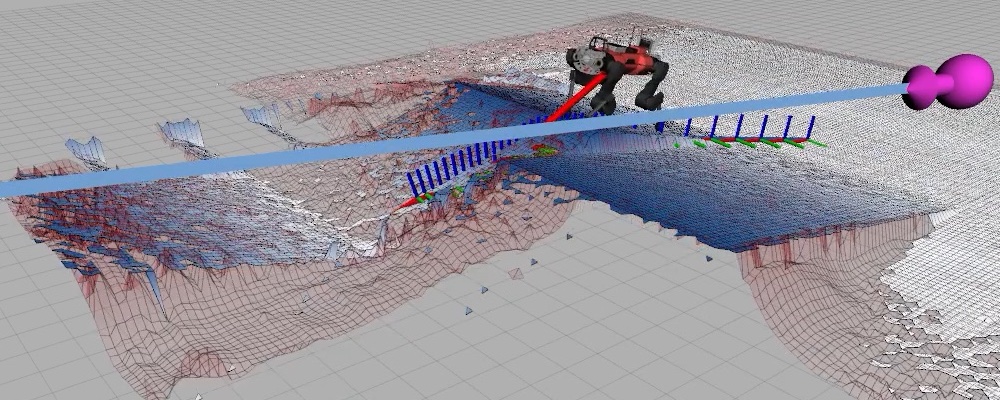}}
  \hfill
  \subfloat{\includegraphics[width=0.495\textwidth]{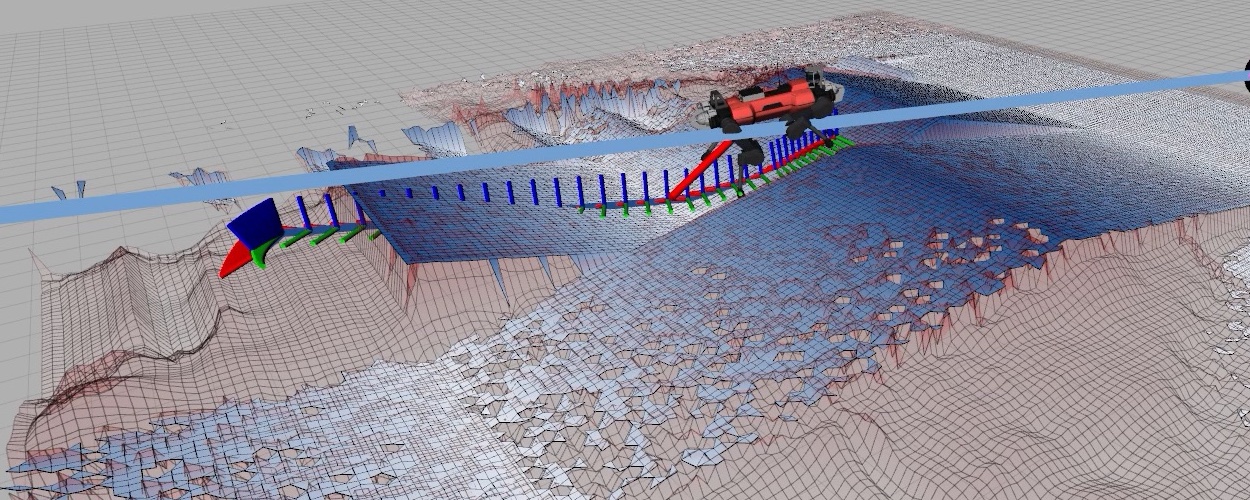}}
  \caption{Samples visualizing motion planning for an ANYmal legged robot in a simulated environment.
  The blue line represents the vector from the start pose (purple arrow) on the right side to the goal on the left side.
  \textbf{First row:} incomplete \acp{DEM} are used for motion planning. \textbf{Second row:} \acp{DEM} reconstructed by a neural network trained with self-supervision are employed for motion planning.
  }
  \label{fig:motion_planning}
  \vspace{-0.6cm}
\end{figure*}

\subsection{Experiments on real-world datasets}\label{sub:experiments_real_world}
We evaluate our proposed self-supervised learning approach on several real-world datasets. 
As complete ground-truth \acp{DEM} are hard to acquire in the real world, we evaluate on the test set of the real-world data with artificial occlusion created using ray casting.
We run experiments using five different random seeds and report a confidence interval with mean and standard deviation.


Evaluating quantitatively on real-world datasets containing artificial occlusion generated with ray casting, we state a decrease of \SI{64}{\percent}, \SI{52}{\percent}, \SI{82}{\percent}, and \SI{57}{\percent} in \ac{MSE} error compared to the best performing baseline approach on each dataset using self-supervised learning for the ANYmal ETH stairs, obstacles course, Gonzen mine~\cite{tranzatto2021cerberus}, and Tenerife lunar analogue datasets as listed in Table~\ref{tab:results_real_world_datasets}.
The $\ell_1$ and \ac{PSNR} error metrics are substantially improved likewise.
When we analyse error distribution of our self-supervised approach, we note that \SI{75}{\percent} of samples have an $\ell_1$ error smaller than \SI{5}{cm} for the ETH stairs, \SI{4}{cm} for the ETH obstacle course, \SI{4}{cm} for the Gonzen mine, and \SI{3}{cm} for the Tenerife dataset.

\subsection{Qualitative comparison}\label{sub:qualitative_comparison}
We visualize qualitative subgrid samples comparing our self-supervised learning approach with the ground-truth and the Telea~\cite{telea2004image} baseline method in Figure~\ref{fig:qualitative_samples}. 
Please note that the ground-truth is already partially occluded and we applied additional artificial occlusion using ray casting to generate our input \ac{DEM}.
In general, we can say that our approach is able to reconstruct the DEM with more details and at a lower error compared to Telea~\cite{telea2004image}.
In particular, our method is showing promising indications in leveraging not only the information contained in the input \ac{DEM}, but also of the geometric constraints concerning line of sight by estimating that if a patch is fully occluded, it probably has a lower elevation than the surrounding area as seen in the reconstruction in the upper right corner of the third row.
Here, we would like to mention one potential drawback in using a self-supervised approach when training on real-world data: as the neural network is trained on data which contains noise, it will also learn to reconstruct similar noise patterns for the occluded areas.

\subsection{Run-time}\label{sub:results_runtime}
We evaluate the run-time of some baselines methods on CPU and the proposed U-Net neural network on CPU and GPU. 
The methods running on CPU had access to one core of a 2.3 GHz Intel Core i9 processor in a Mid-2019 MacBook Pro. We used an NVIDIA TITAN Xp GPU to evaluate the run-time of the neural network. 
The scenario we use to evaluate the run-time is inspired by the ANYmal datasets: We assume occluded DEMs of size 300x300px as our input.
We divide the DEM into 16 subgrids each of size 75x75px and subsequently downsample the subgrids to 64x64px.
We then pass the DEM in two batches each of batch size 8 to the neural network and report the run-time for the processing of the entire full-size DEM. 
The baseline methods Telea~\cite{telea2004image} and Navier-Stokes~\cite{ebrahimi2013navier} can be run at around \SI{50}{Hz} and our proposed U-Net with a sampling rate of \SI{3.5}{Hz} on CPU and \SI{30}{Hz} on GPU for a full-size \ac{DEM} enabling real-time inference on autonomous mobile robots.


\subsection{Preliminary study on motion planning}

As part of a preliminary investigation into using our approach for navigation, we utilize the completed \acp{DEM} as an input for motion planning for the ANYmal legged robot in a simulated environment. The elevation maps are inpainted with a neural network trained on the ETH Stairs dataset. We point out the impressive generalization from the real-world stairs dataset to the simulation environment depicted in Figure~\ref{fig:motion_planning}.
The locomotion~\cite{miki2022wild} and navigation~\cite{wellhausen2021rough} components used in simulation were used by team CERBERUS during the DARPA Subterranean Challenge~\cite{tranzatto2021cerberus}.
First, we execute motion planning using sensor-generated, incomplete \acp{DEM} while not planning any path through areas without elevation information. Currently, this baseline scenario is very common in literature~\cite{wellhausen2021rough, gerdes2020efficient}. 
Our experiments show that this approach cannot plan far-ahead and that the paths are not cost-optimized over the entire map. 
Accordingly, the robot needs to stop and wait frequently until the path re-planning is completed. We show examples for this behaviour in the first row of Figure~\ref{fig:motion_planning}.
On the contrary, paths panned based based on completed \acp{DEM} reach further ahead and only need to be slightly adjusted when the belief of the occluded area is updated. We present this behaviour in the second row of Figure~\ref{fig:motion_planning} and in a video comparison on YouTube\footnote{\url{https://youtu.be/_SAZm7tMMUI}}.

\section{Conclusion}\label{sec:conclusion}
This work proposes a self-supervised learning approach for completing sparse 2.5D terrain maps.
The method leverages artificial occlusion generated with an iterative ray casting algorithm to train a neural network on real-world data with an incomplete ground-truth.
We have evaluated our self-supervised learning approach on a variety of synthetic and real-world datasets.
Evaluating quantitatively on real-world datasets containing artificial occlusion generated with ray casting, we state a decrease of between \SI{52}{\percent} and \SI{82}{\percent} in \ac{MSE} error compared to the respective best performing baseline approach using self-supervised learning based on ray casting.
Qualitative samples show that our proposed method is able to both leverage terrain characteristics but also information inherently encoded in the occlusion masks such as full-filling the geometric constraints of line of sight indicating the advantage compared to traditional baseline methods.
The neural network is able to run with sampling rates of \SI{3.5}{Hz} and \SI{30}{Hz} respectively for CPU and GPU inference.
We utilize the reconstructed \acp{DEM} for motion planning in a proof-of-concept. We observe that the path can be planned further ahead using reconstructed elevation maps compared to planning with incomplete elevation maps. Consequently, the robots needs to stop less frequently and wait less for re-planning to finish.

For future work, we could envision several interesting research opportunities: 
Instead of using ray casting, it could be worthwhile to train a separate \ac{GAN} network to learn realistic occlusion patterns to generate additional artificial occlusion as part of the self-supervised learning.
We are not able to use perceptual and styles losses~\cite{gatys2015neural} for our self-supervised learning approach as the adopted pretrained VGG-16~\cite{simonyan2014very} requires a complete ground-truth. As a replacement, we would like to encourage the addition of a discriminator which could be trained to distinguish between reconstructed and ground-truth \ac{DEM}.
Finally, it is crucial to explore model and data uncertainty estimation methods to predict the variance in the neural network predictions to enable safe and optimal motion planning. The cost function optimized during path planning should be influenced by the uncertainty estimation of each grid cell. 






\bibliographystyle{bibliography/IEEEtran}
\bibliography{IEEEabrv, bibliography/references}





\end{document}